\documentclass{article}
\usepackage{spconf,amsmath,epsfig}
\usepackage[linesnumbered, ruled, vlined]{algorithm2e}
\usepackage{algorithmic}
\usepackage{booktabs}
\usepackage[colorlinks=true, allcolors=blue]{hyperref}
\usepackage{booktabs}
\usepackage[table]{xcolor}
\usepackage{multirow}
\usepackage{multicol}
\usepackage{enumitem}

\title{Analyzing Decades-Long Environmental Changes in Namibia Using Archival Aerial Photography and Deep Learning}

\name{\begin{tabular}{c}
    Girmaw Abebe Tadesse\sthanks{Corresponding author: \textcolor{blue}{gtadesse@microsoft.com}}\textsuperscript{1}, Caleb Robinson\textsuperscript{1}, Gilles Quentin Hacheme\textsuperscript{1}, Akram Zaytar\textsuperscript{1}, Rahul Dodhia\textsuperscript{1},\\Tsering Wangyal Shawa\textsuperscript{2}, Juan M. Lavista Ferres\textsuperscript{1}, 
     Emmanuel H. Kreike\textsuperscript{2}
\end{tabular}}

\address{Microsoft AI for Good Research Lab\textsuperscript{1}, Princeton University\textsuperscript{2}}

\begin{document}
%
\maketitle
\begin{abstract}
This study explores object detection in historical aerial photographs of Namibia to identify long-term environmental changes.
Specifically, we aim to identify key objects -- \textit{Waterholes}, \textit{Omuti homesteads}, and \textit{Big trees} -- around Oshikango in Namibia using sub-meter gray-scale aerial imagery from 1943 and 1972. 
In this work, we propose a workflow for analyzing historical aerial imagery using a deep semantic segmentation model on sparse hand-labels. To this end, we employ a number of strategies including class-weighting, pseudo-labeling and empirical p-value-based filtering to balance skewed and sparse representations of objects in the ground truth data. 
Results demonstrate the benefits of these different training strategies resulting in an average $F_1=0.661$ and $F_1=0.755$ over the three objects of interest for the 1943 and 1972 imagery, respectively.  
We also identified that the average size of Waterhole and Big trees increased while the average size of Omutis decreased between 1943 and 1972 reflecting some of the local effects of the massive post-Second World War economic, agricultural, demographic, and environmental changes. This work also highlights the untapped potential of historical aerial photographs in understanding long-term environmental changes beyond Namibia (and Africa). With the lack of adequate satellite technology in the past, archival aerial photography offers a great alternative to uncover decades-long environmental changes.  
\end{abstract}

\section{Introduction}
\label{sec:intro}
Satellite imagery is a valuable source of data that can shed light on the long-term impacts of climate change~\cite{yang2013role}.
However, until the launch of IKONOS in 1999, commercial satellite imagery with a spatial resolution of $<1\text{m}/\text{pixel}$ was not available. The spatial resolution of older satellite images is insufficient to uncover detailed and long-term changes for specific areas of interest. Moreover, the archive of satellite imagery does not start early enough to analyze changes such as the massive post-Second World War global transformation -- the Landsat-1 satellite was the first that collected continuous imagery over the Earth starting in 1972~\cite{loveland2012landsat}. In contrast, archival aerial photographs -- widely available since the early 20th century (for military observation, mapping and planning) provide a longer temporal coverage bringing the post-Second World War “Second Industrial Revolution” or “Great Acceleration” into focus at sub-meter resolution to monitor subtle changes on the ground in local areas. Massive stocks of historical aerial photos remain underutilized in archives across the globe. 
For example, the US National Archives preserves 35 million historical aerial photos; tens of millions more are found in private and state archives, store rooms and offices in other countries~\cite{kreike2024archival}.

In this work, we aim to utilize archival aerial photos from north-central Namibia, taken in 1943 and 1972, to uncover the decades long changes on the ground predating the introduction of high-resolution satellite imagery. The 1943 aerial photos are assumed to be the first instance where aerial photography technology was systematically
used in capturing the landscape of northern Namibia~\cite{shawa2023creating}.
This region is home to a significant portion of Namibia's population, but it is highly vulnerable to climate changes due to its semi-arid environment. Individual aerial photos were first digitized, geo-referenced and joined into a large orthomosaic for further machine learning (ML) driven analysis as described in~\cite{shawa2023creating}. We particularly focused on identifying Waterholes, Omuti homesteads and Big trees.
\textit{Waterholes} used to be the main source of water for the population in the dry season, which resulted in a dispersed settlement pattern of  \textit{Omuti homesteads} in the past. \textit{Big trees}, e.g., marula and palm trees, were main sources of nutrition~\cite{kreike2013environmental}. 

\begin{figure*}[t]
    \centering
    \includegraphics[width=0.9\linewidth]{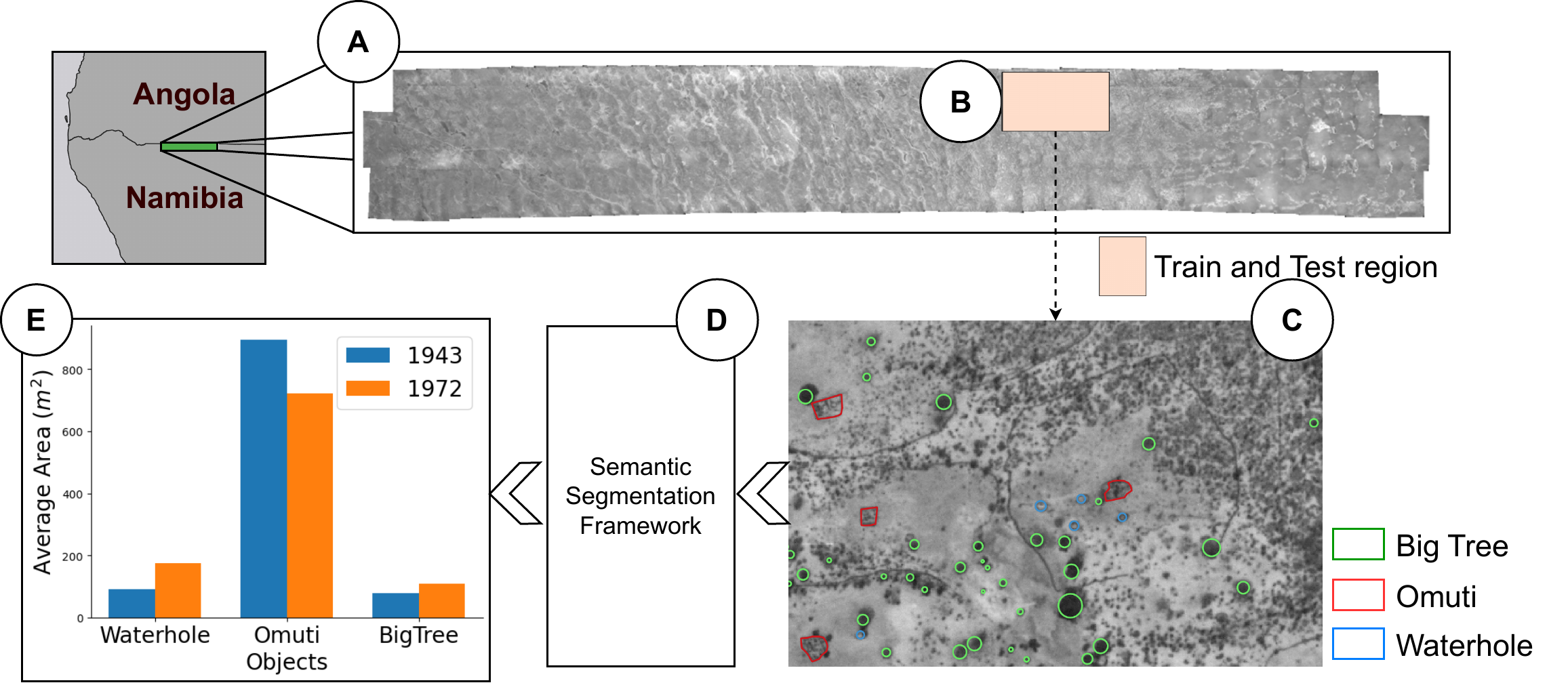}
    \caption{Overview of the proposed approach. Our study focuses on identifying objects of interest from decades-long aerial photos (1943-1972) to study long-term environmental changes in: (A) the Oshikango region ($\approx 5000$ $km^2$) in the North-Central Namibia; (B) a $45$ $km^2$ area in Oshikango region was sparsely annotated and used as train and test region in our framework; (C) representative examples were annotated for the classes: \textit{Big Tree}, \textit{Omuti} and \textit{Waterhole}; (D) a deep learning framework that aims to apply a semantic segmentation on the aerial photos and trained with different strategies; (E) insights are extracted to understand the change between 1943 and 1972.}
   \label{fig:overview}
\end{figure*}

With the encouraging potential of machine learning (ML) algorithms to decipher large collection of data and identify patterns, we employ a deep learning framework that aims to take the digitized aerial photos as input and detected these objects of interest.  Specifically, the framework contains a U-Net-based segmentation model~\cite{ronneberger2015u} with a backbone of a pre-trained ResNet~\cite{he2016deep} network.  To validate the framework, we utilized a sparsely annotated portion of a $45$ $km^2$ area as our train and test region. Once the model was trained, we scaled up the detection stage to identify \textit{Waterholes, Omuti homesteads} and \textit{Big trees} in an area of $\approx 5000$ $km^2$.
In summary, this work offers the following contributions: i.) utilizing aerial photos to identify long-term environmental changes, ii.) a  class weighting strategy that jointly optimizes both the sparsity of annotated objects (classes) and the inter-class imbalance, iii.) empirical p-value based post-processing to plausibly select pseudo-labels from the previous prediction stage for a semi-supervised learning strategy.

The remainder of the paper is organized as follows. Section~\ref{sec:method} presents the methodology, with details on the main contributions. Section~\ref{sec:experiments} describes the experimental setup including the specifics of the datasets used, segmentation model employed and its setting, and evaluation metrics. We present the notable results and follow up discussions in Section~\ref{sec:results}. Finally, Section~\ref{sec:conclusion} concludes the paper with next steps as a future work.

\begin{figure*}[ht]
    \centering
    \includegraphics[width=0.9\linewidth]{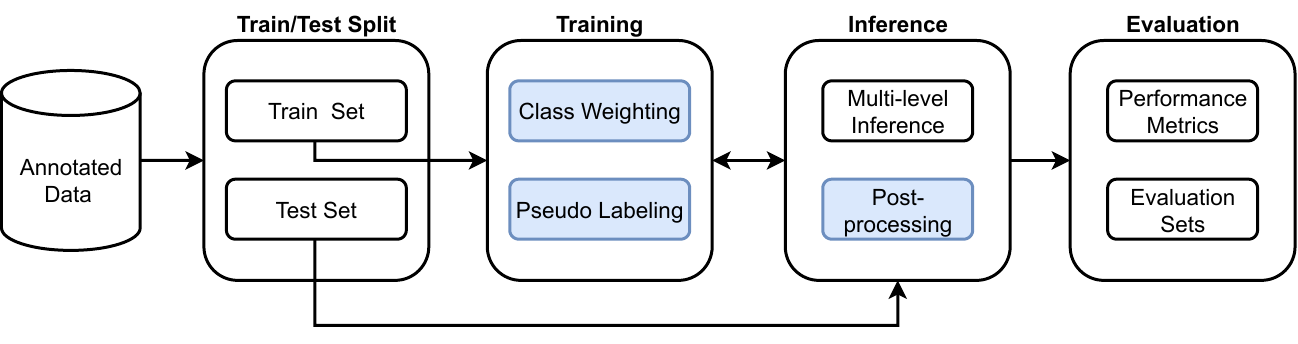}
    \caption{The block-diagram of the proposed segmentation framework, where the highlighted blocks constitute the main contributions. Given a sparse set of annotated data for a $~\approx 45~km^2$ area of the Oshikango region, we, first, split the data spatially to non-overlapping train and test sets. We employed a deep learning model for the segmentation task, which utilizes a \textit{U-Net}~\cite{ronneberger2015u} architecture using  a pre-trained ResNet~\cite{he2016deep} architecture as a back bone. The \textit{Training} step utilizes a \textit{Class Weighting} strategy due to the sparse nature of the annotation, and \textit{Pseudo-labeling} to exploit the originally unannotated part of the train set.  \textit{Inference} is performed at \textit{pixel} and \textit{polygon} levels. The \textit{Evaluation} step adopts segmentation metrics to quantify the performance of the model. Evaluations sets include the test set (with ground truth data) and the whole Oshikango region.}\label{fig:ml_pipeline}
    \label{fig:methodolgoy}
\end{figure*}

\section{Methodology}\label{sec:method}
The overview of our approach is shown in Fig.~\ref{fig:overview}. Given aerial photos from the Oshikango region in Namibia from 1943 and 1972, we aim to detect specific objects of interest: \textit{Big Trees, Omuti homesteads} and \textit{Waterholes} at each of the time stamps to uncover long-term environmental changes. To this end, we employed a pre-processing step that aims to digitize and geo-reference each of the photos and merge them into a large orthomosaic input following the steps in ~\cite{shawa2023creating}. The domain experts annotated a sparse examples of these objects in the subset of the input data ($\approx 45$ $km^2$ ) as a ground truth data for our semantic segmentation framework (shown in Fig.~\ref{fig:ml_pipeline}). Next, we describe the details of the main steps in the framework.

\subsection{Problem Formulation}\label{subsec:problem}
Let $\bf D_t$ represents an orthomosaic  of multiple aerial photos taken in a year, $t$, after each photo is digitized and geo-referenced. The problem is related to evaluating the potential of these aerial photos to quantify the long-term environment changes by detecting a set of objects of interest - $b$: \textit{Big tree}, $o$: \textit{Omuti} and $w$: \textit{Waterhole} at $t=1943$ and $t=1972$. To this end, we employ a deep learning framework to detect these objects at each $\bf D_t$ with a dedicated model, $\bf \Theta_t$. We assume a few examples of these objects are available as polygons  or mask data, $\bf M_t$,  annotated by an experienced expert in the region. Each pixel, $m_i \in \bf M_t$, assumed to be one of the classes: $\mathcal{C} = \{b,o,w, u\}$, where $u$ represents \textit{unknown} or \textit{background} pixels. Due to the  sparse nature of annotation performed in a smaller region, i.e., $|\bf M_t|) << |\bf D_t|)$, where $|\cdot|$ represents dimension,  a key aspect of the framework involves effective usage of $\bf M_t$ where the number of labeled pixels, $N_k$, is quite small compared to the \textit{unknown} pixels, $N_u$. Furthermore, there is a large degree of imbalance  in annotated pixels among classes $b, o$ and $w$. This also poses a critical question on how to utilize the larger number of $u$  pixels in $\bf M_t$  thereby assisting the training process and  enhancing detection performance. $\bf M_t$ is, first, split into train ($\bf M_t^r$) and test ($\bf M_t^e$) splits with no overlapping between the two splits. The model, $\Theta_t$, is trained  using $\bf M_t^r$ and evaluated for both $\bf M_t^r$ and $\bf M_t^e$. We further extend $\bf M_t^r$ by incorporating new masks derived from predicted polygons from previously unannotated regions in $\bf D_t$ as pseudo-labels.

\subsection{Class weighting}\label{subsec:class_weighting}
Class imbalance is a common challenge in geospatial machine learning as it is often resource-demanding to do manual annotations, resulting in a sparse set of annotated regions. This is also partly due to the different observation frequencies of objects of interest. For example, in the related aerial photo Oshikango region in this work, we observed a higher occurrence of \textit{Big trees} compared to \textit{Omuti homesteads}.  In addition, the coverage area of each class may vary (see Table~\ref{tab:class_data_distribution}) resulting imbalanced number of pixels across classes.
Varieties of solutions have been employed to address class imbalance challenges over the years that could be clustered under \textit{re-sampling}~\cite{cui2019class,buda2018systematic} and \textit{re-weighting}~\cite{cui2019class, huang2019deep}. Re-sampling includes sampling minority classes~\cite{cui2019class}, which can lead to overfitting, or under-sampling majority classes~\cite{buda2018systematic}, potentially losing valuable data in cases of extreme imbalance. Data-augmentation also helps to synthetically generate additional samples for minority classes~\cite{zou2018unsupervised}. On the other hand, re-weighting assigns adaptive weights to classes often inversely to the frequency of the class~\cite{cui2019class, huang2019deep}.  Sample-based re-weighting, such as Focal loss, adjusts weights based on individual sample characteristics, targeting well-classified examples and outliers~\cite{lin2017focal, li2019gradient}.

In this work, we propose a simple class weighting strategy that considers both the sparsity of annotated regions) (compared to unannotated regions) and the imbalance of pixels annotated across the classes of interest. Our weighting strategy follows a re-weighting approach that  aims to provide a class weight that is inversely proportional to the ratio of pixels annotated per each class (compared to the remaining classes). Let $N_w$, $N_o$ and $N_b$ be the number of pixels annotated with  \textit{Waterhole}, \textit{Omuti} and \textit{Big tree} classes in the training set, $\bf M_t^r$, respectively. The number of \textit{unlabeled} pixels is denoted by $N_u$. The total number of pixels in $M_t^r$ is $N_k + N_u$, where $N_k=N_w + N_o + N_b$.  Thus, the weight of each class is formulated as follows:
    $\lambda_u = N_k/(N_k + N_u)$,
   $\lambda_w = N_k/N_w$,
    $\lambda_o = N_k/N_o$, and
    $\lambda_b = N_k/N_b$,


\subsection{Pseudo-labeling and Post-processing }\label{subsec:pseudolabel_postprocessing}
To further improve the efficiency of our training steps, we propose to incorporate the weak labels generated from the inference of the model on the previously un-annotated pixels in training -- i.e. a pseudo-labeling based approach~\cite{rizve2020defense,chen2023boosting}. We assume that we have large amounts of unlabeled imagery, however we only have sparse labels (Section~\ref{subsec:problem} ). Once the deep semantic segmentation model,$\Theta_t$, is trained and  model parameters $W_\theta^t$ are obtained, the inference is applied on the training set $\bf M_t^r$, resulting a class prediction probability for each pixel, $m_i \in \bf  M_t^r$. New instances for each class are then recruited from the pseudo-labels to further train the model semi-supervised.

However, deep learning models are known to provide over-confident predictions even in cases where the predicted classes are not correct~\cite{zhang2023survey}, 
which may result in re-training our model with noisy labels. To this end, we propose a post-processing approach that is based on an empirical p-value derived from the features of the predicted polygons, such as area and perimeter. This approach is motivated by recent studies on the robustness of  deep learning frameworks, where similar empirical evaluations were conducted to identify out-of-distribution samples coming from synthesized content~\cite{cintas2022pattern} or adversarial attacks~\cite{kim2023spatially}. This approach also aligns with a growing interest in data-centric research~\cite{oala2023dmlr} that aims to improve model performance by focusing on the data than the model, e.g., by improving the quality of data~\cite{liang2022advances}.

\begin{figure}[htb]

\begin{minipage}[b]{0.48\linewidth}
  \centering
  \includegraphics[width=0.95\linewidth]{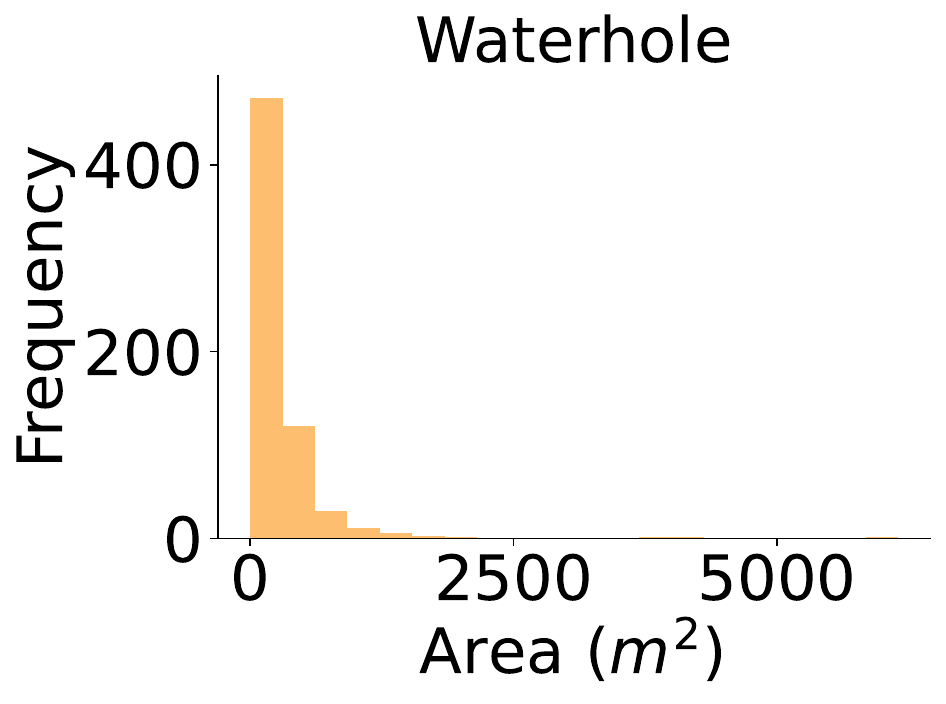}
  \includegraphics[width=0.95\linewidth]{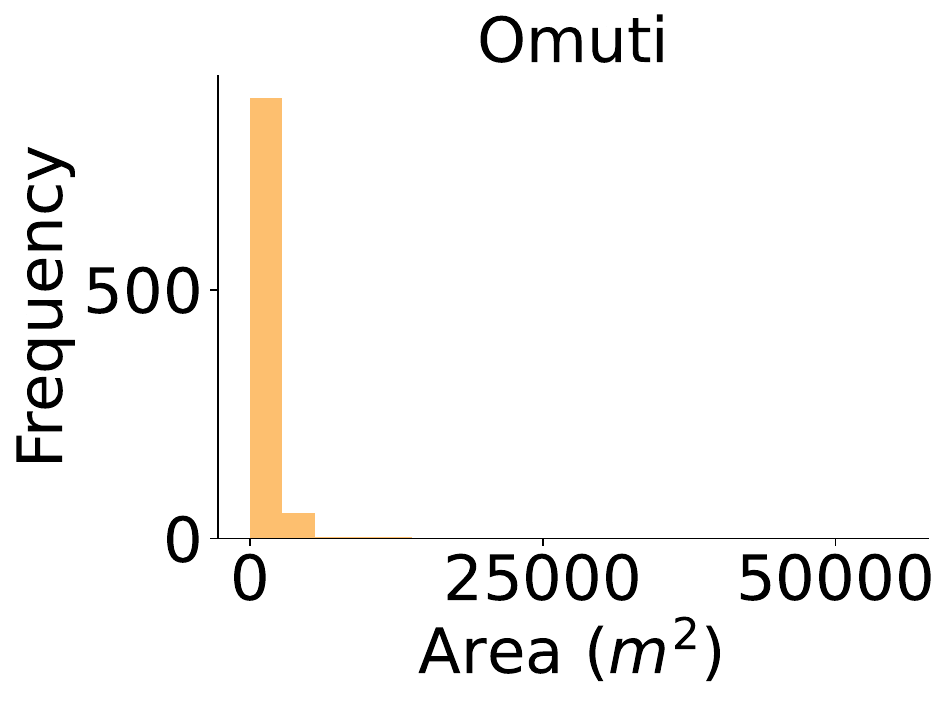}
  \includegraphics[width=0.95\linewidth]{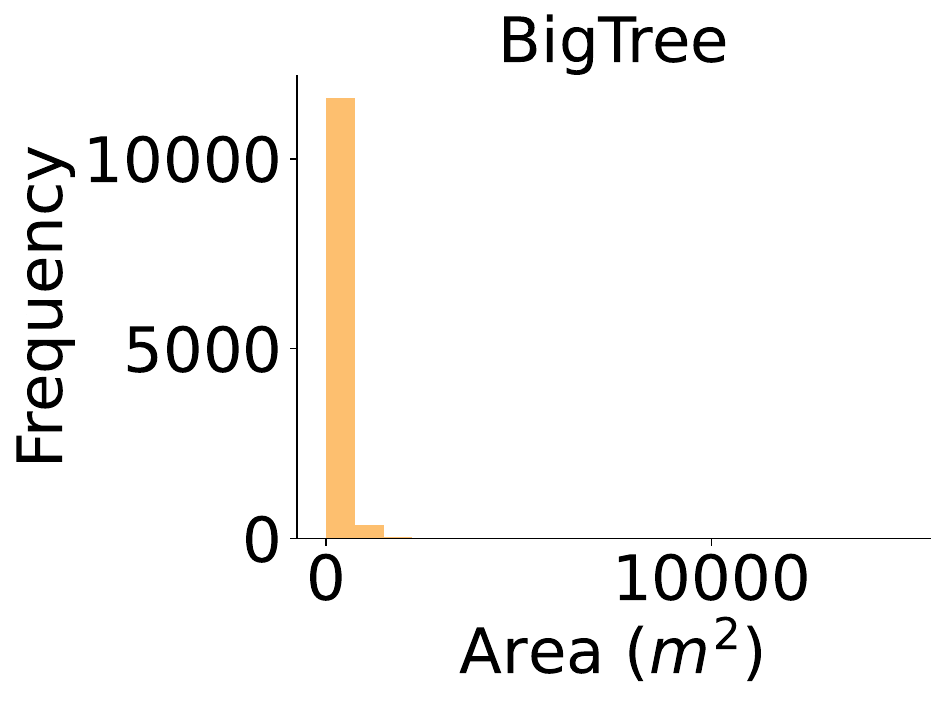}
  \centerline{(a) Area histogram}\medskip
\end{minipage}
\begin{minipage}[b]{.48\linewidth}
  \centering

    \includegraphics[width=0.95\linewidth]{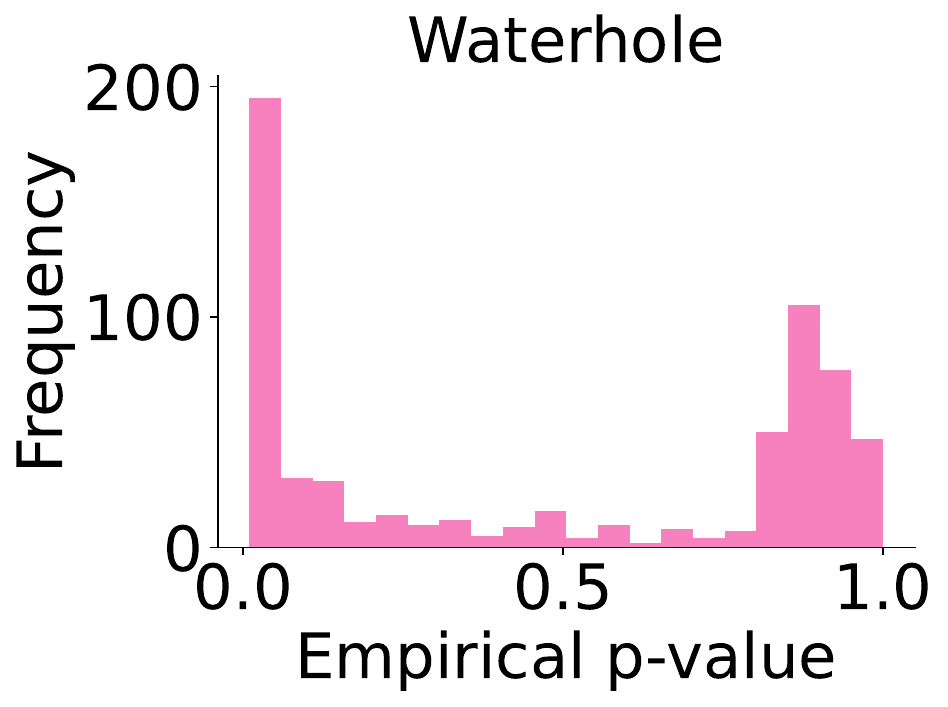}
  \includegraphics[width=0.95\linewidth]{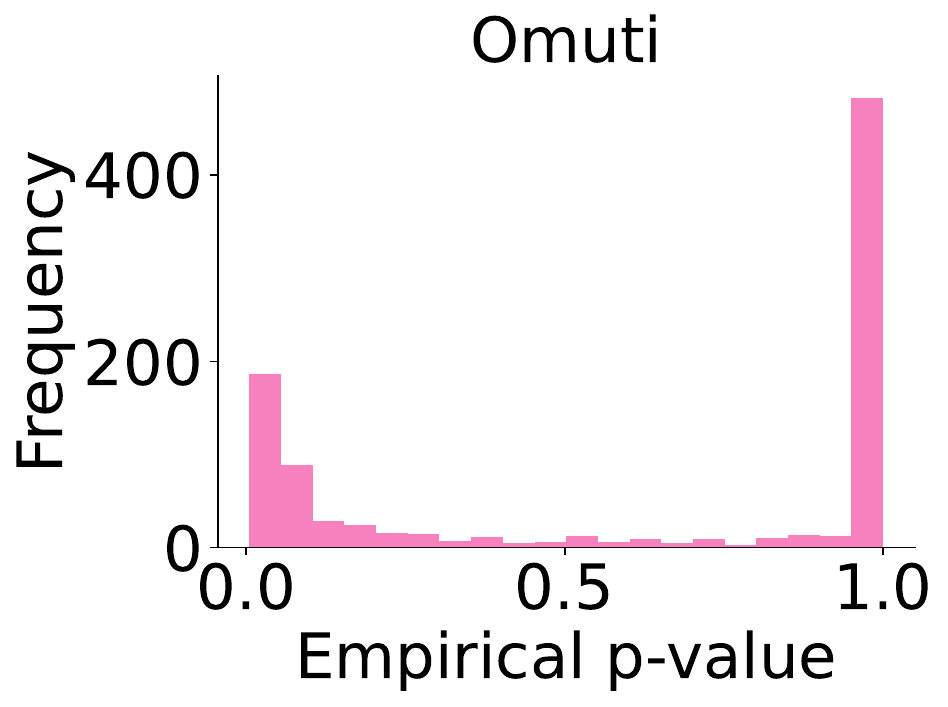}
   \includegraphics[width=0.95\linewidth]{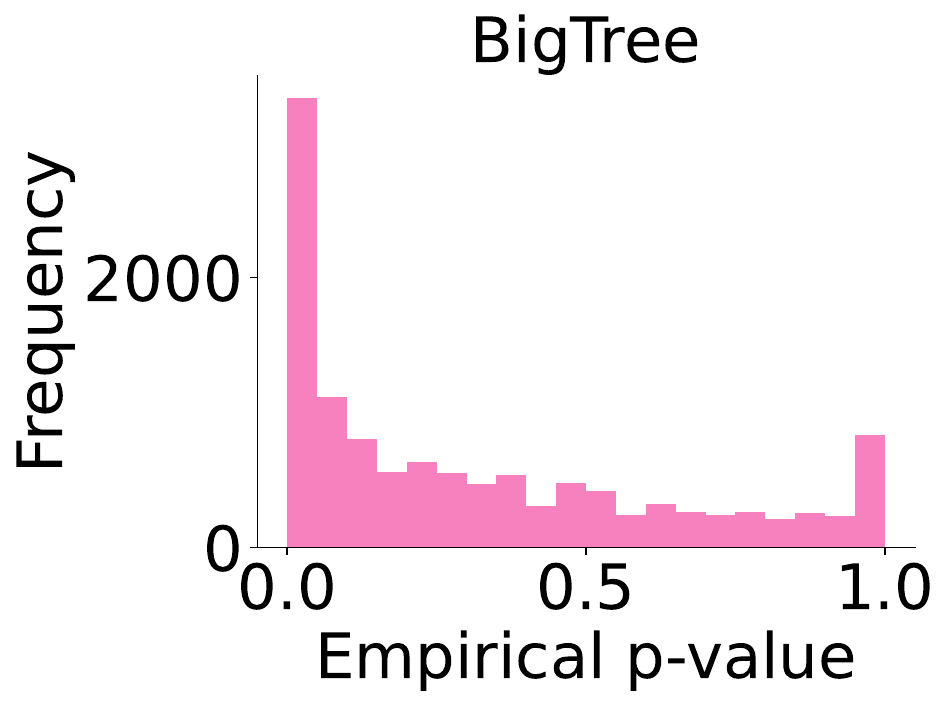}
  \centerline{(b) Histogram of area p-values}\medskip
\end{minipage}
\caption{Visualizations of the areas of predicted polygons from 1943 imagery as (a) histogram  and (b) empirical value compared to the polygons in the training set for each  class. It is clear that the histogram distributions are heavily skewed and the post-processing filter will be are very sensitive of the threshold value. On the other hand, the empirical p-value distributions show more balance and hence less sensitive for a threshold-based post-processing. }
\label{fig:post-processing-examples}
\end{figure}

In this work, we aim to utilize \textit{area} feature and discard predicted polygons with area values that are out-of-distributions from areas of training polygons per each class.   Typical threshold-based filtering could be applied directly on the histogram of the area values. However, it has been found that the distribution is skewed heavily (see Fig.~\ref{fig:post-processing-examples} (a)) and hence a threshold based filtering will be very sensitive to the threshold value. On the other hand, the distributions of empirical p-values (see Fig.~\ref{fig:post-processing-examples} (b)) is relatively less skewed and hence more stable for threshold-based filtering. The pseudo-code of the proposed empirical p-value based post-processing is shown in Algorithm~\ref{alg:post-processing}. Let's assume that we are given the set of annotated training polygons, $P_a^j$, and predicted polygons, $P_p^j$, for each $j^{th}$ class of interest in $\hat{\mathcal{C}}=\{b,o,w\}$. Then we compute the area of each annotated polygon in $P_a^j$, e.g., $A_{an}^j$. Similarly,  we compute the area of each polygon in $P_p^j$, e.g., $A_{pk}^j$. The empirical p-value of each predicted polygon, $E_{pk}^j$, is calculated by counting the number of polygons in $P_a^j$ that are greater than or equal with $A_{pk}^j$. Thus, a predicted polygon with out-of-distribution area will have extreme empirical p-value, i.e., $E_{pk}^j\approx 0$ and  $E_{pk}^j\approx 1$ for predicted polygons with very small area and very large area, compared to the training set, respectively.  Finally,  the predicted polygons that satisfy the empirical-value-based threshold, $e_{th}^j$, are considered as pseudo labels to be used in the follow up recursive training steps.

\begin{algorithm}[t]
\caption{Pseudo-code for the proposed empirical p-value based post-processing that aims to discard predicted polygons that are out-of-distribution from the annotated polygons per each class. }
\label{alg:post-processing}

\SetKwFunction{CountPolygons}{CountPolygons}
\SetKwFunction{ComputeArea}{ComputeArea}
\SetKwFunction{FilterByThreshold}{FilterByThreshold}
\SetKwInOut{Input}{input}
\SetKwInOut{Output}{output}
\Input{Training set: $\bf{M}_t^r$,\\ 
Annotated classes in $\bf{M}_t^r$: $\hat{\mathcal{C}}=C$ \textbackslash $u$,\\
Annotated polygons in $\bf{M}_t^r$: $P_a^j$,\\
Predicted polygons in $\bf{M}_t^r$: $P_p^j$,\\
Filtering threshold: $e_{th}$,\\
}
\Output{Filtered set of predicted polygons in $\bf{M}_t^r$: $\hat{P_p}$} 
\BlankLine
\For{$c_j$ in $\hat{\mathcal{C}}$}{
$N_a^j$ $\leftarrow$ \CountPolygons($P_a^j$)\;
\For{$n$ $\leftarrow$ $1~ \KwTo ~N_a^j$}{
 $A_{an}^j$ $\leftarrow$ \ComputeArea($P_{an}^j$)\;
 }
 
$N_p^j$ $\leftarrow$ \CountPolygons($P_p^j$)\;
\For{$k$ $\leftarrow$ $1~ \KwTo ~N_p^j$}{
$A_{pk}^j$ $\leftarrow$ \ComputeArea($p_{j}$)\;
 $E_{pk}^j$ $\leftarrow$ $\frac{1+\sum_{n=1}^{N_a^j}(A_{an}^j \geq A_{pk}^j)} {N_a^j + 1}$\;
 }
 $\hat{P_p^j}$ $\leftarrow$ \FilterByThreshold($E_p^j,e_{th}$)\;
 }

\Return $\hat{P_p}$
\end{algorithm}

\section{Experimental Setup}\label{sec:experiments}
In this section, we describe the data sources used in this work, along with the distribution of annotations across classes, the deep learning model architecture and hyper-parameters set for our experiments, and performance evaluation metrics.

\subsection{Dataset}
\begin{figure}
     \centering
     \includegraphics[width=0.95\linewidth]{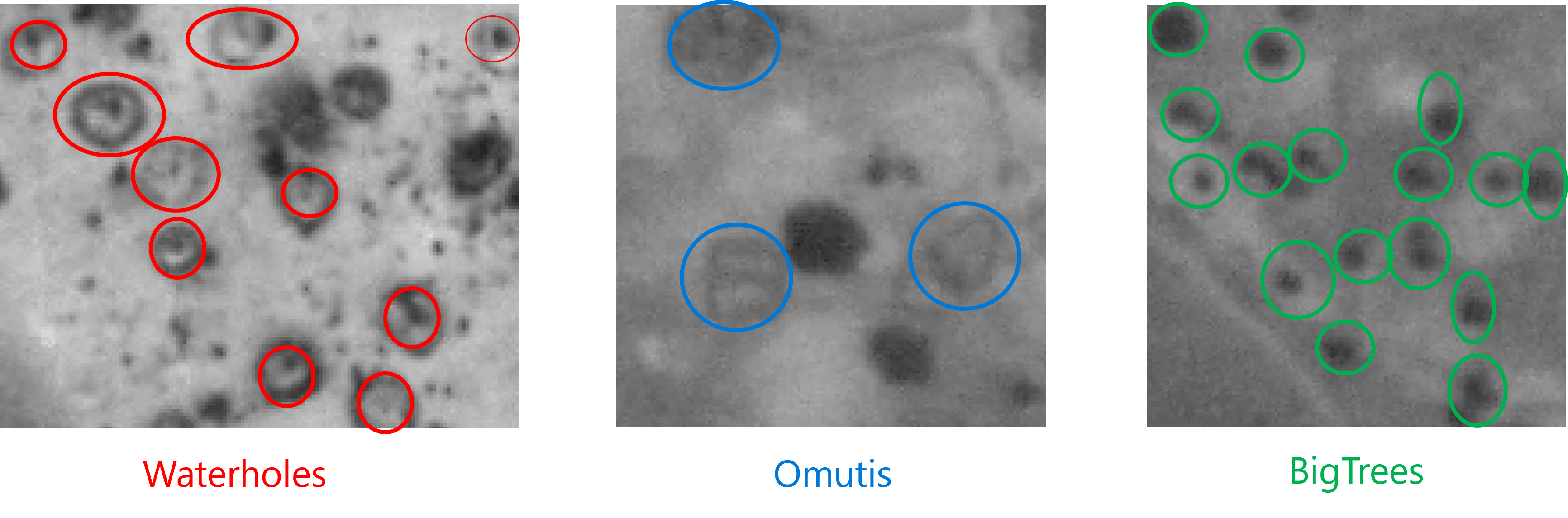}
     \caption{Examples of annotated aerial photos and the three objects of interest, i.e., \textcolor{red}{Waterholes}, \textcolor{blue}{Omuti homesteads} and \textcolor{green}{Big trees}}.
     \label{fig:aerial_photo_example}
 \end{figure}

We have used aerial photos taken in Northern Namibia in the years 1943 and 1972. See Fig.~\ref{fig:aerial_photo_example} for an example of these photos and the types of classes annotated in these photos. Note the annotations were done manually by a domain expert. Table~\ref{tab:class_data_distribution} shows the distributions of annotations in pixels and polygons. The aggregated percentage of annotated pixels is $<1\%$ in 1943 imagery and $<4\%$ in 1972 imagery demonstrating the sparsity of  annotated pixels (regions) compared to the unannotated regions - a typical challenge in geospatial imagery. Furthermore, Table~\ref{tab:class_data_distribution} demonstrates the nature of the imbalanced number of annotated pixels or polygons across classes, e.g., $\approx90\%$ or more of these annotated polygons belong to \textit{Tree} class whereas \textit{Waterhole} constitute only $<4\%$ of the polygons in both 1943 and 1972 images.


\begin{table*}[tb]
\centering
\resizebox{\linewidth}{!}{%
\begin{tabular}{@{}lcccccccccccc@{}}
\toprule
 \multirow{2}{*}{\textbf{Class}} & \multicolumn{2}{c}{\bf 1943 Pixels} & \multicolumn{4}{c}{\bf 1943 Polygons} & \multicolumn{2}{c}{\bf 1972 Pixels} & \multicolumn{4}{c}{\bf 1972 Polygons} \\ \cmidrule(l){2-3} \cmidrule(l){4-7} \cmidrule(l){8-9} \cmidrule(l){10-13} 
 & \# & \% & \# & \% & Sum Area($m^2$) & Mean Area ($m^2$) & \# & \% & \# & \% & Sum Area($m^2$) & Mean Area ($m^2$) \\ \midrule
Waterhole & 39776 & 0.04 & 103 & 3.44 & 14809 & 143.77 & 52397 & 0.11 & 273 & 1.97 & 47850 & 175.27 \\
Omuti & 483095 & 0.42 & 205 & 6.85 & 190892 & 931.18 & 350400 & 0.74 & 482 & 3.48 & 348398 & 722.82 \\
BigTree & 580251 & 0.51 & 2685 & 89.71 & 230052 & 85.68 & 1410979 & 2.99 & 13088 & 94.55 & 1410616 & 107.78 \\ \midrule
Total & 1103122 & 0.97 & 2993 & 100.00 & 435753 & 145.59 & 1813776 & 3.84 & 13843 & 100.00 & 1806864 & 130.53 \\ \bottomrule
\end{tabular}%
}
\caption{Distribution of annotated pixels, polygons and their corresponding areas from $1943$ and $1972$ aerial photographs across the three objects of interest for this study: \textit{Waterhole}, \textit{Omuti} and \textit{Big tree}. Both the number (\#)  and percentage (\%) of annotated pixels and polygons per each object are given. In addition, the average and total sum of area on the ground annotated for each object are provided. }\label{tab:class_data_distribution}
\end{table*}

\subsection{Model Selection and Set-up}
We have employed a U-Net-based~\cite{ronneberger2015u} semantic segmentation deep learning framework, using a pre-trained ResNet-50~\cite{he2016deep} architecture as a backbone architecture for each of the 1943 and 1972 aerial photos. We have also employed a $70\%-30\%$ train-test split of the annotated regions. We also employed a cross-entropy loss and a learning rate of $0.001$. The batch size and maximum epochs were set to $64$ and $50$, respectively. Note that the pixels with no annotation are treated as background class, and weighted accordingly so as not to affect the optimization significantly.

\subsection{Multi-level Inference and Performance Evaluation Metrics}
Inference is performed at each pixel level in the test set, which can also be aggregated to polygon-level inference. Thus, the evaluation metrics are also computed corresponding to the level of inference. Generally, we employed Accuracy, Precision, Recall, and $F_1$ score to evaluate how well each class's annotated pixels (regions) were detected during inference.  
All the four metrics represent detection performance based on \emph{true positive (tp)}, \emph{true negative (tn)}, \emph{false positive (fp)} and \emph{false negative (fn)} values. For pixel-level performance metrics, 
true positive is when the class pixel is correctly identified; true negative is when the pixels associated with the remaining classes are correctly identified as negative; false positive is the case when pixels corresponding to the remaining classes are incorrectly detected as the class pixel, and false negative refers to the case when class pixels are incorrectly detected as the remaining class pixels. For polygon-level performance metrics, tp, tn, fp, fn are computed from a threshold-based overlapping  of regions, e.g., $5\%$, between the predicted and ground truth polygons. Note that we have not computed the evaluation metrics for the background {$u$} class as it could  still be a real background  class or any of the classes but left unlabeled during annotation. 

\section{Results and Discussion}\label{sec:results}
\subsection{Performance of different training strategies}\label{subsect:training_strategies}

 \begin{table}[!t]
    \centering
    \resizebox{0.99\linewidth}{!}{
    \begin{tabular}{llccc}
    \toprule
        \textbf{Year} & \textbf{Training Strategy} & \textbf{Precision} & \textbf{Recall} & $\mathbf{F_1}$  \\ \midrule
        \multirow{4}{*}{1943} & Baseline & 0.421 & 0.261 & 0.295  \\ 
        ~ & ``+'' Class Weighting & 0.273 & 0.656 & 0.381  \\ 
        ~ & ``+'' Pseudo Labeling & \bf{0.436} & 0.816 & \bf{0.549}  \\ 
        ~ & ``+'' Post-processed Pseudo Labeling & 0.306 & \bf{0.866} & 0.434  \\ \midrule
        \multirow{4}{*}{1972} & Baseline & 0.662 & 0.663 & 0.629  \\ 
        ~ & ``+'' Class Weighting & 0.375 & \bf{0.809} & 0.495  \\ 
        ~ & ``+'' Pseudo Labeling & 0.677 & 0.720 & 0.697  \\ 
        ~ & ``+'' Post-processed Pseudo Labeling & \bf{0.688} & 0.735 & \bf{0.706} \\ \bottomrule
    \end{tabular}
    }
    \caption{Detection results averaged across the Waterhole, Omuti and Big Tree classes in 1943 and 1972 when different training strategies are employed in our semantic segmentation framework. \textbf{Bold} values represent the highest-performing training strategy per each  imagery and performance metric.  Class weighting aims to provide a weighting factor per class that is inversely proportional to the observation frequency. Pseudo-labeling aims to utilize predicted samples that were not in the annotation set back to training - a key strategy to use the majority the previously unlabeled training regions. Furthermore, pseudo-labeling could be applied with or without post-processing of the predicted samples. Our post-processing steps adopts empirical p-value based threshold ($e_{th}$) computed from the area of predicted polygons compared to the training polygons. We found $e_{th}=0.5$ STD for 1943 imagery and $e_{th}=1.0$ STD for 1972 imagery worked better.}\label{table:raw_averages}
\end{table}
Table~\ref{table:raw_averages}
shows the results derived from different training strategies employed in our semantic segmentation framework across two imagery timestamps: 1943 and 192. Compared to $F_1=0.549$ in 1943 imagery, we achieved a higher $F_1=0.706$ on the 1972 imagery. This is partly due to the higher number of examples from 1972 imagery to train its model (see Table~\ref{tab:class_data_distribution}). Furthermore, our different training strategies, i.e., class weighting, pseudo labeling and post-processed pseudo-labeling, outperformed the Baseline that does not include any of these strategies. Particularly, the class weighting strategy alone improved the Recall values from $0.261$ to  $0.656$ in 1943 imagery and from $0.663$ to $0.809$ in 1972 imagery by effectively weighting the cross-entropy loss by the inverse of the observation of each class. Pseudo-labeling that aims to utilize high-confident predictions in a semi-supervised learning fashion is also shown to further improve the Precision (by reducing the false positives) and then the $F_1$ score in both imagery sources. Additional difference between 1943 and 1972 images involve the impact of using pseudo-labels after empirical p-value based post-processing (i.e., \textit{Post-processed Pseudo Labeling}). Since the ground truth data of 1943 imagery suffers from very few number of training samples per class (i.e., $<1\%$ of the imagery is annotated), discarding the predicted polygons based on a threshold did not result in an improved performance. On the other hand, filtering the pseudo-labels before recursive training improved all the metrics in 1972 imagery, resulting in the highest $F_1=0.706$.

\subsection{Impact of post-processing on evaluation metrics}\label{subsec:post_processing_on_evaluation}
Tables~\ref{table:f1_averages} demonstrates the need of filtering predicted polygons as a part of our empirical area p-value based post-processing even for evaluating the performance metrics. The highest average $F_1$ score across the  three classes is achieved in 1943 ($F_1=0.661$) imagery using a p-value threshold of $e_{th}=0.5$ STD. Similarly, the post-processing has improved the average $F_1$ score from $0.706$ to $0.755$ using an p-value threshold of $e_{th}=1.0$ STD, which is partly due to a larger and more balanced set of ground truth data and hence does not require a strong threshold that would discard more polygons.

\begin{table}[t]
    \centering
    \resizebox{0.99\linewidth}{!}{
    \begin{tabular}{llccc}
    \toprule
    \multirow{2}{*}{\textbf{Year}} & \multirow{2}{*}{\textbf{Training Strategy}} & & \multicolumn{2}{c}{\textbf{Post-processing ($e_{th}$)}} \\
    \cmidrule(r){4-5}
        & & \textbf{Raw} &  $\mathbf{1.0}$ \textbf{STD} &$\mathbf{0.5}$ \textbf{STD}  \\ \midrule
        \multirow{4}{*}{1943} & Baseline & 0.295 & 0.284 & 0.255  \\ 
        ~ & ``+'' Class Weighting & 0.381 & 0.455 & 0.466  \\ 
        ~ & ``+'' Pseudo Labeling & 0.549 & 0.620 &\bf{0.661}  \\ 
        ~ & ``+'' Post-processed Pseudo Labeling & 0.434 & 0.473 & 0.643  \\ \midrule
        \multirow{4}{*}{1972} & Baseline & 0.629 & 0.643 & 0.609  \\ 
        ~ & ``+'' Class Weighting & 0.495 & 0.584 & 0.610  \\ 
        ~ & ``+'' Pseudo Labeling & 0.697 & 0.745 & {0.729}  \\ 
        ~ & ``+'' Post-processed Pseudo Labeling & 0.706 & \bf{0.755} & 0.725 \\ \bottomrule
    \end{tabular}
    }
    \caption{Average $F_1$ results across the two-time stamps and training strategies when different post-processing thresholds ($e_{th}$) is applied to filter the pseudo-labels before evaluating the metrics. \textbf{Bold} values represent the highest $F_1$ score achieved for each timestamp. Note that post-processing with a threshold ($e_{th}=0.5$ STD), which discards more polygons compared to $e_{th}=1.0$ STD,  consistently performed better for 1943 imagery across different training strategies. On the other hand, 1972 imagery does not require such as a substantial filtering threshold as it works best for $e_th=1.0$ STD, partly due to a larger and more balanced set of ground truth data.}\label{table:f1_averages}
\end{table}

\subsection{Analyzing false positives}\label{subsec:analysis_false_positives}
\begin{figure}[t]

\begin{minipage}[b]{.99\linewidth}
  \centering
    \includegraphics[width=0.99\linewidth]{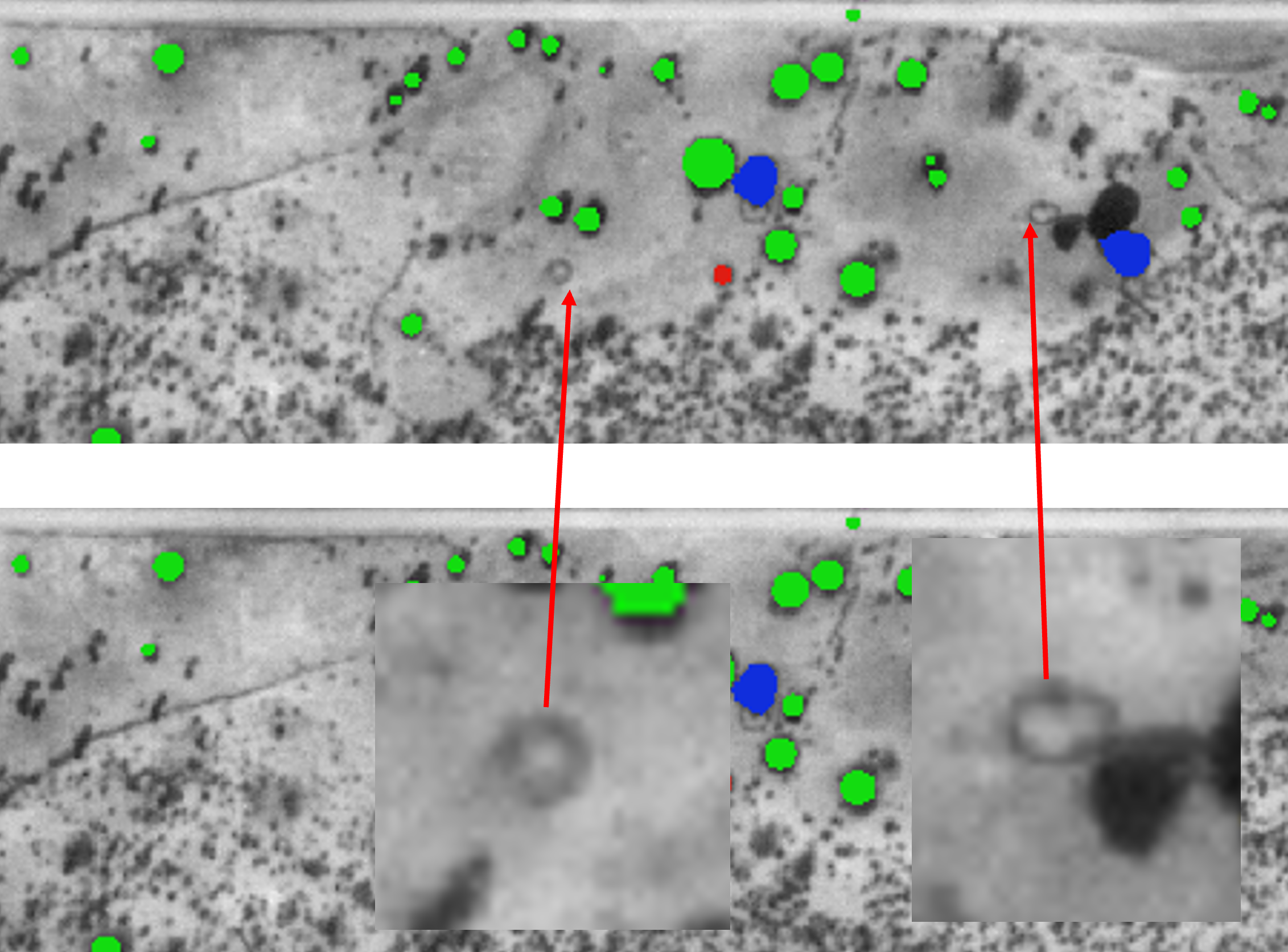}
  \centerline{(a) Ground truth polygons}\medskip
\end{minipage}

\begin{minipage}[b]{.99\linewidth}
  \centering
    \includegraphics[width=0.99\linewidth]{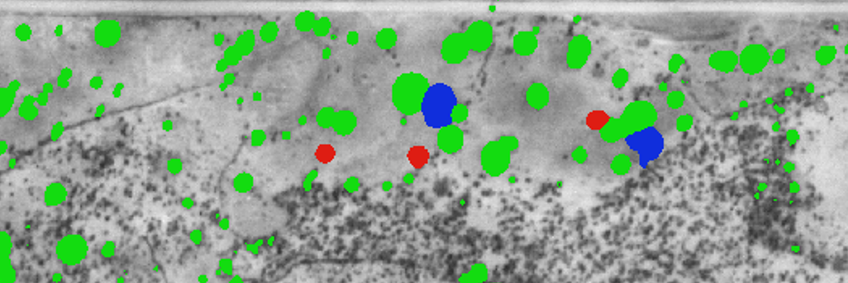}
  \centerline{(b) Predicted polygons}\medskip
\end{minipage}

\caption{False positive could be previously unlabeled objects during the annotation (a) is the ground truth data where two objects were not labeled during annotation, and the arrows show the zoomed version of these objects for better visualization, and (b) these two objects were then detected as Waterholes during inference time (marked with \textcolor{red}{dots}). Note that \textit{Big trees} and \textit{Omuti homesteads} are marked with \textcolor{green}{green} and \textcolor{blue}{blue} markers, respectively. 
}
\label{fig:false-positive-examples}
\end{figure}

Among the metrics employed to evaluate our detection performance, Recall values are found to be consistently higher than Precision values for both imagery sources (see  Table~\ref{table:raw_averages}). This is partly due to the higher observation of false positives compared to false negatives. Further analysis demonstrates that all false positives are not actually falsely identified objects of interest but previously unlabeled objects. Figure~\ref{fig:false-positive-examples} shows such an instance, where   two objects were shown unlabeled in the ground truth polygons in Figure~\ref{fig:false-positive-examples} (a). But these two objects are detected as Waterholes during inference time, thereby suggesting the framework could also help to discover objects of interest that were not labeled during annotation though they might still be evaluated as false positives. This further motivates the need of pseudo-labeling in our framework that aimed to utilize such objects that were left unlabeled during annotation but later found out to be objects of interest with high confidence.


\begin{figure}[htb]
    \centering
    \includegraphics[width=0.7\linewidth]{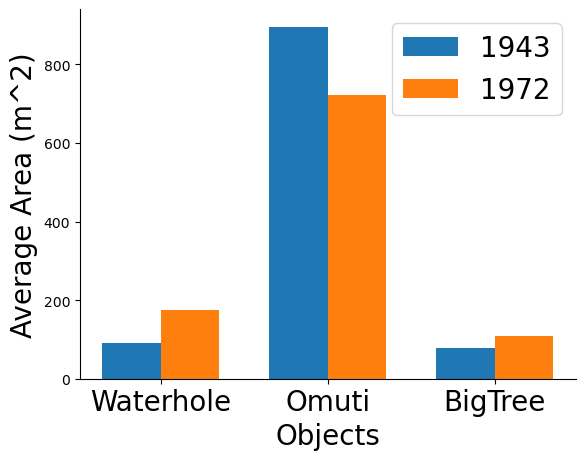}
    \caption{Environmental changes observed across Waterhole, Omuti and Big Tree objects when their average areas ($m^2$) are compared between 1943 and 1972. The average sizes of Waterhole and Big Tree increased whereas the size of Omuti got smaller in those years.}
    \label{fig:changes_1943_72}
\end{figure}

\subsection{Changes between 1943 and 1972 and scalability}\label{subsec:changes}
Moreover, our analysis facilitates understanding of past events with less/limited data, and provides added quantitative and qualitative details on historic reports (see Fig.~\ref{fig:changes_1943_72}). For example, the number and location of Waterholes and homesteads confirm the sharing of Waterholes by neighbors, low-yielding reports of these water holes, and increased population density. Furthermore, class-based changes (e.g., Waterholes) in number, area, coverage, locations and/or proximity to other objects of interest reveal further insights on the changes that took place between 1943 and 1972. 

Furthermore, we have utilized the trained semantic segmentation model in a larger area covering  $\approx 5000$ $km^2$. See Fig.~\ref{fig:overview}(B) to visualize the scale of a region compared to the relatively smaller annotated region ($\approx 45$ $km^2$) used as a ground truth. Such large scale implementation of the framework helps to benefit domain experts by reducing the resource necessary to make exhaustive annotation, and generate large scale insights.

\section{Conclusions and Future Work}\label{sec:conclusion}
Understanding long-term environmental changes requires the use of old and remotely sensed images such as satellite imagery. However, the resolutions of satellite images were not at sub-meter level a few decades back.
Old aerial photos, on the other hand, satisfy these requirements though they are often stored unused in archives and museums across the world. In this work, we aim to demonstrate the capabilities for understanding long-term environmental changes in Namibia using aerial photos taken during 1943 and 1972 using deep learning to detect the following objects: \textit{Waterholes}, \textit{Omuti homesteads} and \textit{Big trees}.
To this end, we employed a deep semantic segmentation framework that includes U-Net~\cite{ronneberger2015u} model  with a pre-trained ResNet-50~\cite{he2016deep} architecture as its backbone. To address the challenges associated with the sparseness of annotated regions and imbalance among classes, we proposed a class weighting strategy followed with a pseudo labeling step that aims to utilize predicted polygons. The pseudo-labels were further filtered using an empirical p-value based post-processing step.
The results demonstrate the capabilities of aerial photos to understand long-term environmental changes by detecting those classes with encouraging performance. 
Thus, efforts need to be accelerated to digitize and analyze  them to better understand long-term environmental and socio-demographic changes. This work highlighted that aerial photos provide a promising alternative to study environmental changes prior to 1990s as there was no adequate satellite technology to capture images with $<1m/pixel$ resolution. 

Future work aims to investigate further use cases where lower detection performance metrics, both in precision and recall, were observed.  In addition, we aim to  scale up the  validation of the proposed approach beyond the use case in  Namibia. Deploying under-utilized archival aerial photographs is, potentially, a promising alternative to validate current understanding of the past and uncover new insights, which is critical to ensure sustainability in Africa,  where  climate change poses a significant and disproportionate risk compared to its emission.


\bibliographystyle{IEEEbib}
\bibliography{refs}

\end{document}